\documentclass[8pt,twoside,a4paper]{article}
\usepackage{amsmath}
\usepackage{amsfonts}
\usepackage{natbib}
\usepackage{amssymb}
\usepackage{hyperref}
\usepackage{graphicx}
\usepackage{stfloats}
\usepackage{algorithm}
\usepackage{algorithmic}
\usepackage{enumerate}
\usepackage{amsmath}
\usepackage{graphics}
\usepackage{epsfig}
\usepackage{graphicx}
\usepackage{caption2}
\usepackage{subfigure}
\usepackage{float}
\title{Reinforcement Learning with Goal-Distance Gradient}  
\author{Kai Jiang\footnote{jiangkaiff@std.uestc.edu.cn}, XiaoLong Qin\footnote{qxlxajh@163.com}}   
\date{}
\usepackage[a4paper,left=35mm,right=35mm,top=25mm,bottom=25mm]{geometry}

\begin{document}
\maketitle

\begin{abstract}
Reinforcement learning usually uses the feedback rewards of environmental to train agents. But the rewards in the actual environment are sparse, and even some environments will not rewards. Most of the current methods are difficult to get good performance in sparse reward or non-reward environments. Although using shaped rewards is effective when solving sparse reward tasks, it is limited to specific problems and learning is also susceptible to local optima. We propose a model-free method that does not rely on environmental rewards to solve the problem of sparse rewards in the general environment. Our method use the minimum number of transitions between states as the distance to replace the rewards of environmental, and proposes a goal-distance gradient to achieve policy improvement. We also introduce a bridge point planning method based on the characteristics of our method to improve exploration efficiency, thereby solving more complex tasks. Experiments show that our method performs better on sparse reward and local optimal problems in complex environments than previous work.
\end{abstract}

\section{Introduction}
Reinforcement learning (RL) is widely used to train an agent, such as robots, to perform a task by feedback rewards in environment. For example, train an agent to play FPS and Card-Based games~\cite{song2019playing,liu2019playing}, to defeat a champion at the game of Go~\cite{silver2016mastering}, to overtake human scores in 49 Atari games~\cite{guo2016deep}, as well as learn control manipulator arm screw a cap onto a bottle~\cite{levine2016end}, building blocks~\cite{nair2018visual}. Among the above tasks, perhaps the most central idea of RL is value functions $V(s)$ that represents overall feedback rewards in any state~\cite{sutton1998introduction}. The agent is trained to optimize the value function which caches knowledge of reward in order to learn to perform a single task. However, agents are required to perform multi-goal tasks in various environments where many rewards are sparse, such as address a variety of cooperative multi-agent problems~\cite{fu2019deep}. How can we design a method that can perform multi-goal tasks well in a sparse reward or non-reward environment?

\par We consider these problems in a human way of thinking. In fact, when dealing with complex tasks, humans set goals for themselves and keep approaching them. Therefore, the agent is supposed to set and achieve new goals during self-supervised process. In order to ensure the agent can interact with the environment to know how to approach the goal in training, we must first choose a universal distance function. General value functions $V_g(s)$~\cite{sutton2011horde} represent the rewards of any state $s$ in achieving a potential given goal $g$. In the general RL, the value function is represented by a function approximator $V(s,\theta)$, such as a neural network with parameters $\theta$. The function approximator learns the value of the observed state through the structure in the state space, and expands to the unobserved value. The goals are usually set to what the agent can achieve, and on this basis, the goal space usually has the same structure as the state space. Therefore, the idea of value function approximation can be extended to both states $s$ and goals $g$ by using general value function approximator $V(s,g,\theta)$~\cite{schaul2015universal}. If the value function is related to distance, such as reward is to use the negative Mahalanobis distance in the latent space~\cite{nair2018visual}, the smaller the distance, the bigger the reward. We can also use a general function approximation $D(s,g,\theta)$ to represent the distance function.

\par In order to solve the tasks in sparse reward or even non-reward environments, we consider whether the reward can be represented by the distance that exists in any environment. But it is quite difficult to train a function that can accurately evaluate the distance between states directly from the raw signals provided by the environment. For instance, in most visual tasks, the meanings of value function represented by pixel-wise Euclidean distance are not relevant to the meanings of value function in actual states~\cite{ponomarenko2015image,zhang2018unreasonable}. Therefore, we propose to address this challenges by using the distance function $D(s, g)$ to represent the minimum transition step from state $s$ to goal $g$. In any task, the agent needs to transfer to a new state by constantly choosing actions until the goal is reached. Therefore, in any sparse reward or even no reward environment, the distance function can effectively train agents to complete tasks.

\par In this paper, our main contribution is a general skill technique that perform several different tasks in sparse reward or even non-reward environments. In order to obtain effective training results, we modified the previous framework of standard RL algorithm, abandoned the action-value function, and propose a gradient method to improve policy based on distance. Although the usual \emph{Reward shaping} ~\cite{mataric1994reward,ng1999policy} makes learning vulnerable to local optimization, our method can make agents reach the optimal goals. In addition, our method also combines the bridge point theory in SoRB~\cite{eysenbach2019search}, which makes our method can gain better experience in unexpected tasks.

\par Although there are many methods of reinforcement learning that can be used to solve problems in environments. But current methods must be based on sufficient data and training to accomplish specific tasks. Once the environment changes, the previously obtained models will no longer be applicable. Only by overcoming these barrier can reinforcement learning be used in more real-world environments instead of staying in the simulator. So we must let our agents learn to analyze problems, not just judge based on past experience. Our approach provides an idea for agents to analyze problems and split complex problems into multiple simple problems.


\section{Background}
\subsection{Reinforcement Learning}
Reinforcement learning is the control agent interaction with the environment to get the maximum reward. We model this problem as a \emph{Markov decision process}(MDP), and consider this MDP defined by a set of \emph{state space} $\mathcal{S}$, a set of \emph{action space} $\mathcal{A}$, a reward function $r:\mathcal{S}\times\mathcal{A}\rightarrow \mathbb{R}$, an \emph{initial state distribution} with density $p(s_{1})$, and transition probabilities $p(s_{t+1}|s_{t},a_{t})$. The agent's action is defined by a policy $\pi_{\theta}$ with parameters $\theta$. The goal of policy is to find the parameters $\theta$ that maximize the expected sum of future rewards from the start state, denoted by the performance objective $J(\pi)=\mathbb{E}[R^{\gamma}|\pi]$. The expected sum of future rewards is called a \emph{return}: $R_{t}^{\gamma}=\Sigma_{k=t}^{\infty}\gamma^{k-t}r(s_{k},a_{k})$ where $0<\gamma<1$. We can write the performance objective as an expectation,

\begin{equation}
\begin{aligned}
J(\pi_{\theta })&=\int _{S}\rho^{\pi }(s)\int _{A}\pi _{\theta }(s,a)r(s,a)dads\\
&=\mathbb{E}_{s\sim \rho ^{\pi},a\sim \pi _{\theta }}[r(s,a)].
\end{aligned}
\end{equation}
\par Value functions are defined to be the expected sum of future rewards: $V^{\pi}(s)=\mathbb{E}[R^{\gamma}|S=s;\pi]$ and $Q^{\pi}(s,a)=\mathbb{E}[R^{\gamma}|S=s,A=a;\pi]$. These satisfies the following equation called the Bellman equation,
\begin{equation}
\begin{aligned}
V(s_{t})&=\mathbb{E}_{s\sim \rho ^{(\cdot |s,a)},a \sim \pi _{\theta }}[r(s_{t},a_{t})+\gamma V(s_{t+1})]\\
Q(s_{t},a_{t})&=\mathbb{E}_{s\sim \rho ^{(\cdot |s,a)},a \sim \pi _{\theta }}[r(s_{t},a_{t})+\gamma \underset{a\in \mathcal{A}}{max} Q(s_{t+1},a_{t+1})].
\end{aligned}
\end{equation}

\par The actor-critic is a widely used architecture based on the policy gradient theorem~\cite{sutton1998introduction,Peters2008Natural,Degris2012Model}. The actor-critic consists of two eponymous components. An actor adjusts the parameters $\theta$ of the stochastic policy $\pi_{\theta}(s)$. Instead of the unknown true action-value function $Q^{\pi}(s, a)$, an action-value function $Q^{\omega}(s, a)$ is used, with parameter vector $\omega$. A critic estimates the action-value function $Q^{\omega}(s, a) \approx Q^{\pi}(s, a)$ using an appropriate policy evaluation algorithm such as temporal-difference learning.

\subsection{Deep Deterministic Policy Gradient}
Deep Deterministic Policy Gradients(DDPG)~\cite{lillicrap2016continuous} is an off-policy actor-critic algorithm for continuous action spaces. The DDPG mainly includes two parts \emph{actor} and \emph{critic}. The actor is primarily responsible for a deterministic goal policy $\mu$, and the role of the critic is to approximate the action-value function $Q$ that helps the actor learns the policy. Compared with ordinary stochastic policy gradients, DDPG uses a deterministic policy gradient. The generalised policy iteration~\cite{sutton1998introduction} is commonly used in most model-free reinforcement learning algorithms. Use temporal-difference learning~\cite{bhatnagar2007incremental,Degris12linearoff-policy,Peters2008Natural} or Monte-Carlo evaluation to estimate the action-value function $Q^{\mu}(s,a)$. The policy improvement method is a greedy maximisation of the estimated action-value function, $\mu ^{k+1} = arg\underset{a}{max}Q^{\mu ^{k}}(s,a)$.
\subsection{Hindsight Experience Replay}
Tasks with multiple different goals and sparse rewards have always been a huge challenge in reinforcement learning. For the challenge of sparse rewards, the standard solution is to introduce a new informative reward function that can guide the agent to the goal, e.g $r_g(s,g)=-\left\| s-g \right\|_2$. While such shape rewards can solve some problems, it is still difficult to apply to more complex problems. Multiple goals tasks require more training samples and more efficient samples than single goal tasks from an intuitive perspective. Hindsight Experience Replay(HER)~\cite{andrychowicz2017hindsight} present a technique which effective learning of samples from sparse reward environment. HER not only improves the sample efficiency, but also makes it possible to learn sparse reward signals. The method is based on training universal policies~\cite{schaul2015universal} that takes both the current state and the goal state as inputs. For any trajectory $s_{0},s_{1},s_{2},......,s_{t},...,g$, the most central idea of HER is store the transition $(s_{t},a_{t},s_{t+1},g^{*})$ in the replay buffer, and the $g^{*}$ is not only with the original goal $g$ but also with a subset of other goals. 

\section{Goal Distance Gradient}
To realize the use of distance instead of rewards in reinforcement learning, the following two points must be considered. Firstly, in order to make distance replace reward, it means that reward function $V(s)$ and action-value function $Q(s,a)$ will be replaced by distance function $D(s,g)$. How should the distance function be defined and estimated? Can the previous method of evaluating the action-value function also estimate the distance function? Secondly, Without an action-value function, how can we use the distance function to improve the policy? The distance function $D(s,g)$ cannot provide an effective gradient for the policy to improve. We next describe the estimation of distance function in Section 3.1, and the method of Goal Distance Gradient in Section 3.2.

\subsection{Estimate the Distance Function by TD}
The distance function $D(s,g)$ is used to represent the minimum number of transitions from the state $s$ to the goal $g$. Compared with the value function $V(s)$, the distance function has a clear directionality $s\rightarrow g$. But in fact, the $V(s;g')$ also hides a goal $g'$ that can get the maximum cumulative reward when it is reached. $D(s,g)$ and $V(s,g)$ are equal if we set the feedback reward obtained at each step to 1. It means that each step is a transfer, that is how many transfers have been made from $s$ to $g$ or how many rewards have been accumulated. In this case, we can use the temporal-difference learning evaluation to estimate the distance function, such as Sarsa update~\cite{sutton1998introduction} is used by critic to estimate the action-value function in the on-policy deterministic actor-critic algorithm,
\begin{equation}
\begin{aligned}
\delta_t =r_t + \gamma Q(s_{t+1},a_{t+1})-Q(s_t,a_t)
\end{aligned}
\end{equation}
and Q-learning update is used by critic to estimate the action-value function in the off-policy deterministic actor-critic algorithm,
\begin{equation}
\begin{aligned}
\delta_t =r_t + \gamma Q(s_{t+1},\mu(s_{t+1}))-Q(s_t,a_t).
\end{aligned}
\end{equation}
\par Therefore, we only need to replace the reward $r_t$ with the distance $d_t$, and then we can evaluate the distance function without considering off-policy or on-policy. We can define $d_t$ as the number of transfers of $s_t \rightarrow s_{t+1}$,
\begin{equation}
\begin{aligned}
\delta_t &=d_t + D(s_{t+1},g)-D(s_t,g).
\end{aligned}
\end{equation}
\par But there is still a very important problem here, $d_t$ is always a positive number, so as the iteration progresses, the distance function may become larger and larger, and constantly deviate from the correct estimation. We need to set a fixed transition value as the distance benchmark between identical states. We denote $s_g$ is the state after reaching the goal $g$, and the number of transfers $d_{s_g}$ is 0,
\begin{equation}
\begin{aligned}
D(s_t,g)&=d_t+D(s_{t+1},g)d_t\\
D(s_g,g)&=0.
\end{aligned}
\end{equation}
\par The number of transfers is recorded as 0 only when agent remains in its current state without any transfers. That is, the distance between the same states $s$ is defined as $D(s,s)=0$. Therefore, how to keep agent close to the goal $g$ from the start $s$ is equivalent to how to make $D(s,g)$ function close to 0.
\subsection{Gradients of Distance Policies}
Policy Gradient algorithm is often used in continuous action space, and improves policy by the global maximisation at every step. In deterministic policy algorithm, a simple and efficient way to improve policy is through the gradient of action-value function $Q(s,a)$, rather than globally maximising. For each state $s$, the $\theta$ of policy parameters are updated by the negative gradient $\bigtriangledown_{\theta } Q(s,\mu_{\theta}(s))$, and make $\mu_{\theta}(s)$ to output the action $a^*$ to maximize $Q(s,a^*)$. However, the distance function $D(s,g)$ cannot provide gradient for updating the parameters $\theta$ of $\mu_{\theta}$ like action-value function $Q(s,a)$. Therefore, we propose a new policy improvement method based on goal-distance gradient.

\par We define deterministic policy $a=\pi(s;\theta)$ and a deterministic model $s'=f(s,a)$. The form of the deterministic Bellman equation for the action-value function is $Q(s,a)=r(s,a)+\gamma Q(f(s,a),\mu(f(s,a)))$. So, the relationship between action-value function $Q(s,a)$ and value function $V(s)$ is as follows:
\begin{equation}
\begin{aligned}
Q(s,a)&=r(s,a)+\gamma V(s')\\
&=r(s,a)+\gamma V(f(s,a)).	\label{con:relbetQandV}
\end{aligned}
\end{equation}
\par Use value function $V(s)$ instead of action-value function $Q(s,a)$ for policy improvement:
\begin{equation}
\begin{aligned}
\mu(s) &= arg\underset{a \in A}{max} Q(s,a)\\
&= arg\underset{a \in A}{max} (r(s,a)+\gamma V(f(s,a)))\\
&\approx arg\underset{a \in A}{max} V(f(s,a)). \label{con:argmaxV}
\end{aligned}
\end{equation}
\par Although the value function itself cannot provide gradient for the policy to improve, through the relationship with the action-value function, the policy is improved indirectly. The form of Distance Bellman equation is $D(s,g)=d+{D}'(s',g)$ where $d$ refers to the distance or times of transition. The agent aims to obtain a policy which minimizes the distance between the next state and the goal. Reference Equation \ref{con:relbetQandV} and \ref{con:argmaxV}, there are the following formulas:
\begin{equation}
\begin{aligned}
D(s,g)&=d+{D}(s',g)\\
&=d+{D}(f(s,a),g)\\
&=d+{D}(f(s,\mu(s,g)),g)
\end{aligned}
\end{equation}

\begin{equation}
\begin{aligned}
\mu(s,g)&= arg\underset{a \in A}{min} {D}(s',g)\\
&= arg\underset{a \in A}{min} {D}(f(s,a),g)
\end{aligned}
\end{equation}
\par So, the policy improvement method is to use the gradient $\bigtriangledown_{\theta } D(f(s,\mu_{\theta}(s,g)),g)$ to minimize the distance function $D(s,g)$. The improved policy $\mu^{*}(s,g)$ can output an action $a$ that makes the distance $D(s',g)$ from next state $s'$ to the goal $g$ sufficiently small.When $D(s',g)$ is close enough to 0, agent can achieve the goal $g$.

\subsection{Algorithms}
We summarize the reinforcement learning algorithm of the Goal Distance Gradient (GDG) in Algorithm 1. Our main idea is to make agent perceive the distance of the whole environment, which is to estimate the times of transitions required to reach different states. By setting random goal, estimation of the number of agent transfers from different states to the goal by the the temporal-difference method, and then train it by distance function to move in the direction of decreasing the number of transfers, so as to reach the goal. At the beginning of each episode, we judge whether to find a bridge point that can connect the start to the goal and make the start to the goal closer according to a fixed probability. Than, we use a simple exploration policy to collect enough samples for training and stored in the replay buffer. Finally, we train our policy model based on samples randomly sampled from the replay buffer and finetune the parameters of model.
\begin{algorithm}[tb]
	\caption{Goal Distance Gradient(GDG)}
	\begin{algorithmic}[1]
		\STATE Initialize critic network $D(s,g)$, actor $\mu(s,g)$ and $f(s,a)$ with weights $\theta^{D}$, $\theta^{\mu}$ and $\theta^{f}$
		\STATE Initialize target network ${D}'$ with weight $\theta^{D'} \leftarrow \theta^{D}$
		\STATE Initialize replay buffer $R$ and soft replace rate $\tau$
		\STATE Initialize bridge point search exploit rate $\varepsilon$
		\FOR{episode = 1,M}
			\STATE Sample a goal $g$ and an initial state $s0$.	
			\IF{$random(0,1) < \varepsilon$}
				\STATE Search a bridge point
			\ENDIF
			\FOR{$t = 0, T-1$}
				\STATE Get an action $a_t = \mu(s_t,g)+$ noise
				\STATE Execute $a_t$ and observe a new state $s_{t+1}$
				\STATE Store the transition$(s_{t},a_{t},s_{t+1},g,d)$ in $R$
			\ENDFOR
			\FOR{$i = 0, T-1$}
			\STATE Sample a random minibatch from $R_i$
			\STATE Set $x_{i}=d_{i}+D'(s_{i+1},g_{i})$ and $y_{i}=f(s_{i},a_{i})$
			\STATE Update network by minimizing the loss:
		 	\[
			\begin{aligned}
			L^{D}&=\frac{1}{N}\sum_{i}(x_{i}-D(s_{i},g_{i}))^{2}\\
			L^{f}&=\frac{1}{N}\sum_{i}(y_{i}-s_{i+1})^{2}
			\end{aligned}
			\]
			\STATE Update the actor policy:
			 \[
			\triangledown_{\theta^{\mu}}J\approx\frac{1}{N}\sum_{i}\triangledown_{\theta^{\mu}}D(f(s,\mu(s,g)),g)|_{s_i,g_i}
			\]
			\STATE Update the target networks:
			 \[
			 \theta^{D'} \leftarrow \tau \theta^{D} + (1-\tau) \theta^{D'}
			\]
			\ENDFOR
		\ENDFOR
		
	\end{algorithmic}
\end{algorithm}

\par We summarize the method of searching for bridge points in Algorithm 2. In many complex tasks, it is often difficult to collect good experiences for learning because of insufficient exploration. In order to explore and collect more good experience, we can't be limited by the experience we have gained. Therefore, we should get out of the dilemma of thinking and look for new knowledge that has not yet been discovered. When the agent has learned a policy to reach the goal from the start, we can not give up to find out whether there is a better policy. We use the estimated distance function to find out whether there is any state that can make the distance from the start to the goal shorter than the known distance. If there is such a specific state, it shows that there are better policy for agents to learn. We can set this state as a temporary goal, and then collect the experience that can further improve the policy. We use the bridge point to represent the temporary goal that connecting the start and the goal. In Sec.4.2 We introduced in detail how to apply bridge point in our method.
\begin{algorithm}[tb]
	\caption{Search a bridge point}
	\begin{algorithmic}[1]
		\WHILE{Bridge point not found}
			\STATE $bridge \leftarrow Search()$
			\STATE $d_{sg} = D(start, goal)$
			\STATE $d_{sb} = D(start, bridge)$
			\STATE $d_{bg} = D(bridge, goal)$
			\IF{$d_{sg} > d_{sb} + d_{bg}$}
				\STATE Stop searching
			\ELSIF{Number of searches exceeded}
				\STATE Stop searching
			\ELSE
				\STATE Continue searching
			\ENDIF
		\ENDWHILE 
	\end{algorithmic}
\end{algorithm}

\section{Experiments}
Our experiments were designed to address and answer the following questions: 1) Whether the goal-distance gradient is effective for policy improvement, 2) How does the GDG method with bridge planning perform in complex tasks, 3) Does local optima affect GDG in single goal Tasks.

\subsection{Can goal-distance gradients improve policy?}
The DDPG~\cite{lillicrap2016continuous} is a classic algorithm in deterministic policy reinforcement learning algorithms. The key of the DDPG is to find an action that can reach a state of as large a reward as possible after execution. The key to our algorithm is to find an action that can reach the goal the fastest. If the reward for each step is set to -1 and the reward for reaching the goal is set to 0, the value function in DDPG is the same as the distance function in our method have the same meaning. If the combination of start $s$ and goal $g$ is regarded as the state, then $D(s,g) = -V(s\|g)$$\footnote{$\|$ denotes concatenation}$. In the above case, the only difference between our method and DDPG is how to find an ideal action.

\par In order to exclude the influence of other factors, we found the environment of the 7-DOF fetch robotics arm~\cite{andrychowicz2017hindsight}, in which the distance from the start to the end is computable. The states of the start and the goal are represented by the coordinates in the Euclidean space, and the Euclidean distance can be calculated directly. 

\par In this experiment, we consider the distance from the binary representation state $s$ to goal $g$ of form $d(s,g)=\left\| s-g \right\|_2$. Therefore, we can directly use the forms $D(s,g)=d(s,g)$ and $V(s\|g)=-d(s,g)$ to represent the distance function in our method and the value function in the DDPG. Therefore, we can skip the distance evaluation step and proceed directly to the policy improvement phase. In theory, our method should be consistent with the performance of DDPG in this experiment. From Fig.1 it is clear that both methods can easily accomplish this task, and the convergence effect and the final result of the two methods are consistent, and it means that our method is feasible and effective.
\begin{figure}[h]
\centering
\includegraphics[width=8cm]{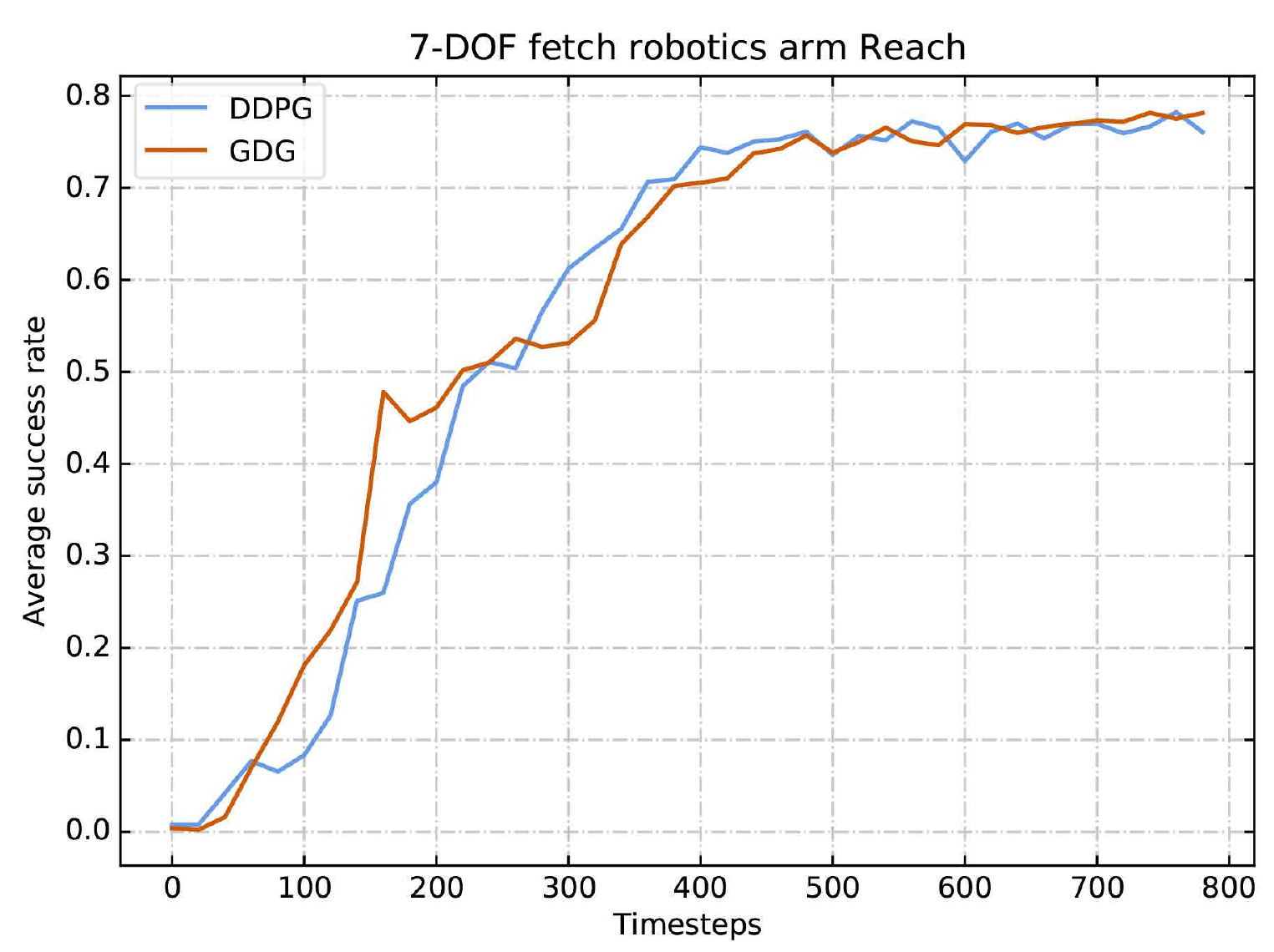}
\label{2}
\caption{Success rates for GDG and DDPG in the 7-DOF fetch robotics arm to reach the goal.
}
\end{figure}
\subsection{Application of bridge point in GDG}
This experiments will illustrate that accurate distances estimates are crucial to the success of our method. ~\cite{eysenbach2019search} define $d_{sg}(s, g)$ as the expected number of steps to reach $s$ from $g$ under the optimal policy. But, this method is not suitable for all environments. In the real-world environment, most of the environments can not estimate the optimal distance from $s$ to $g$ in advance. Therefore, we try to use our distance function $D(s,g)$ instead of $d_{sg}(s, g)$ in SoRB to estimate the distance from $s$ to $g$ in advance. So as to help search bridge point to establish the connection between the start $s$ and the goal $g$ to better complete the task. 

\par In training, it is difficult for agents to learn how to get around obstacles and reach the goal. Like Figure 2-a, it usually moves directly in the direction of goal and hits an obstacle in FourRooms environment. It is more arduous to reach the opposite side of an obstacle, if the obstacle is wider. Because for a seemingly immediate goal, the agent must stay away from it before reaching it. It's complicated for agents to understand why it need to stay away from the goals before it get close to goal. If we can find a bridge point $p_{b}$ that connects start and goal like a bridge, we can guide the agent to reach $p_{b}$ first, then from $p_{b}$ to goal. Because the task of an agent is no longer to stay away from the goal and then reach the goal, but to reach bridge point $p_{b}$ and then reach the goal. It's easier for agent to understand how to accomplish two simple tasks in succession. As shown in Figure 2-b, We find a bridge point $B$ that agent can reach, and Figure 2-c shows that $B$ can also reach the goal. Figure 2-d shows that with the help of the bridge point B, the agent can complete the task from the start to the goal. However, more bridge points may be needed in the actual task to complete the connection between the start and the goal. 
\begin{figure}[htb]
\centering
{\includegraphics[width=8cm]{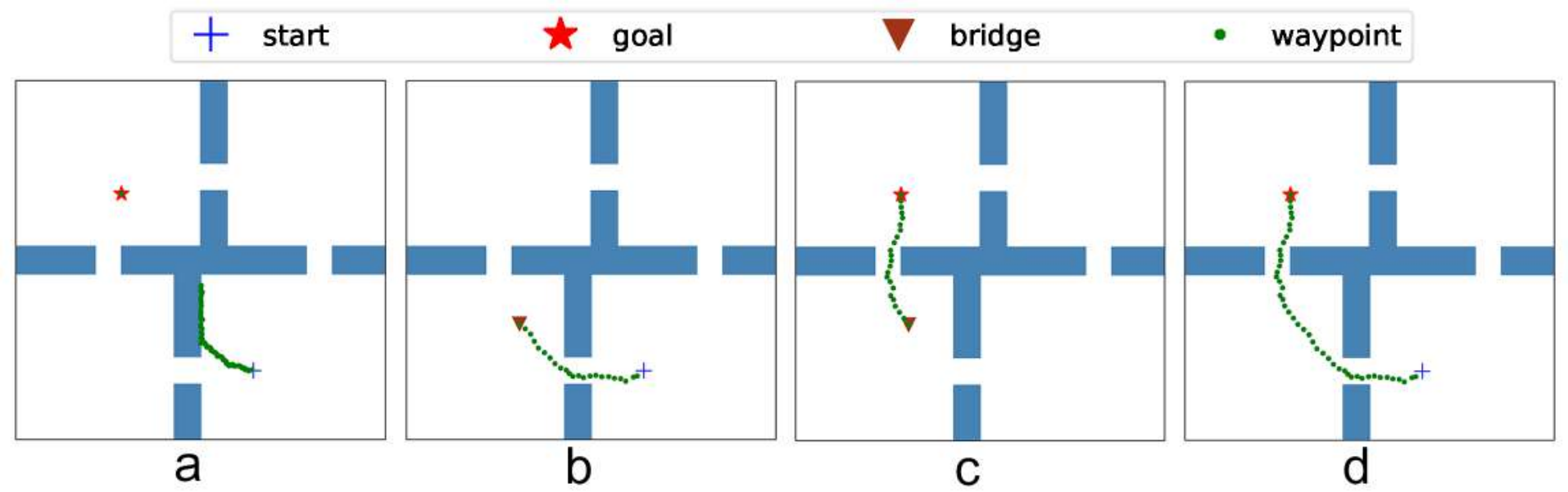}
\label{3}}
\caption{Bridge point establishes the connection between the start and the goal: (a)No bridge points connect the start and the goal, (b)Find a bridge point that can connect to the start, (c)Find a bridge point that can connect to the goal, (d)With the help of the bridge point, the path from the start to the goal was found.}
\end{figure}

\subsection{Performance comparison in complex environments}
Now we test the performance of our method in a more intricate environment, illustrated in Figure 5. We use similar methods to build an environment similar to city streets with more obstacles, more tortuous paths and a larger scope. This environment compared with that simple FourRooms map, the maximum distance (steps) from the start to the goal is increased from 120 to 240, and more obstacles need to be bypassed by agents. We found that is a hard task for the agent to get around multiple obstacles to reach the goal. In FourRooms environment, the agent can reach the goal only by bypassing two obstacles at most. But in city environment, agents need to bypass up to nine obstacles to reach the furthest goal. 

\par In this experiment, we should compare with the SoRB~\cite{eysenbach2019search} algorithm, but the distance it uses is obtained directly from the environment in advance. The distance used in our method is later learned from the environment. Therefore we cannot compare our method with SoRB in this experiment. So, we evaluated five methods: Goal-Distance Gradient, Goal-Distance Gradient with Bridging Planning, Deep Deterministic Policy Gradients(DDPG)~\cite{lillicrap2016continuous}, Hindsight Experience Replay(HER)~\cite{andrychowicz2017hindsight} and stochastic method. We use the same goal sampling distribution when comparing each method.

\par During the training of the above method, each method was tested 200 times for every 20,000 episodes of training, and the average success rate of the results was calculated. From the results in Figure 3, it can be seen that the success rate of DDPG and HER in the late training period is basically maintained at about 0.2 and cannot continue to increase. Although the success rate of my method is still unable to continue to increase in the late training period, it basically remains around 0.5. The success rate of the GDG with bridging planning has been rising in waves, and finally maintained at 0.8.
\begin{figure}[h]
\centering
\includegraphics[width=7cm]{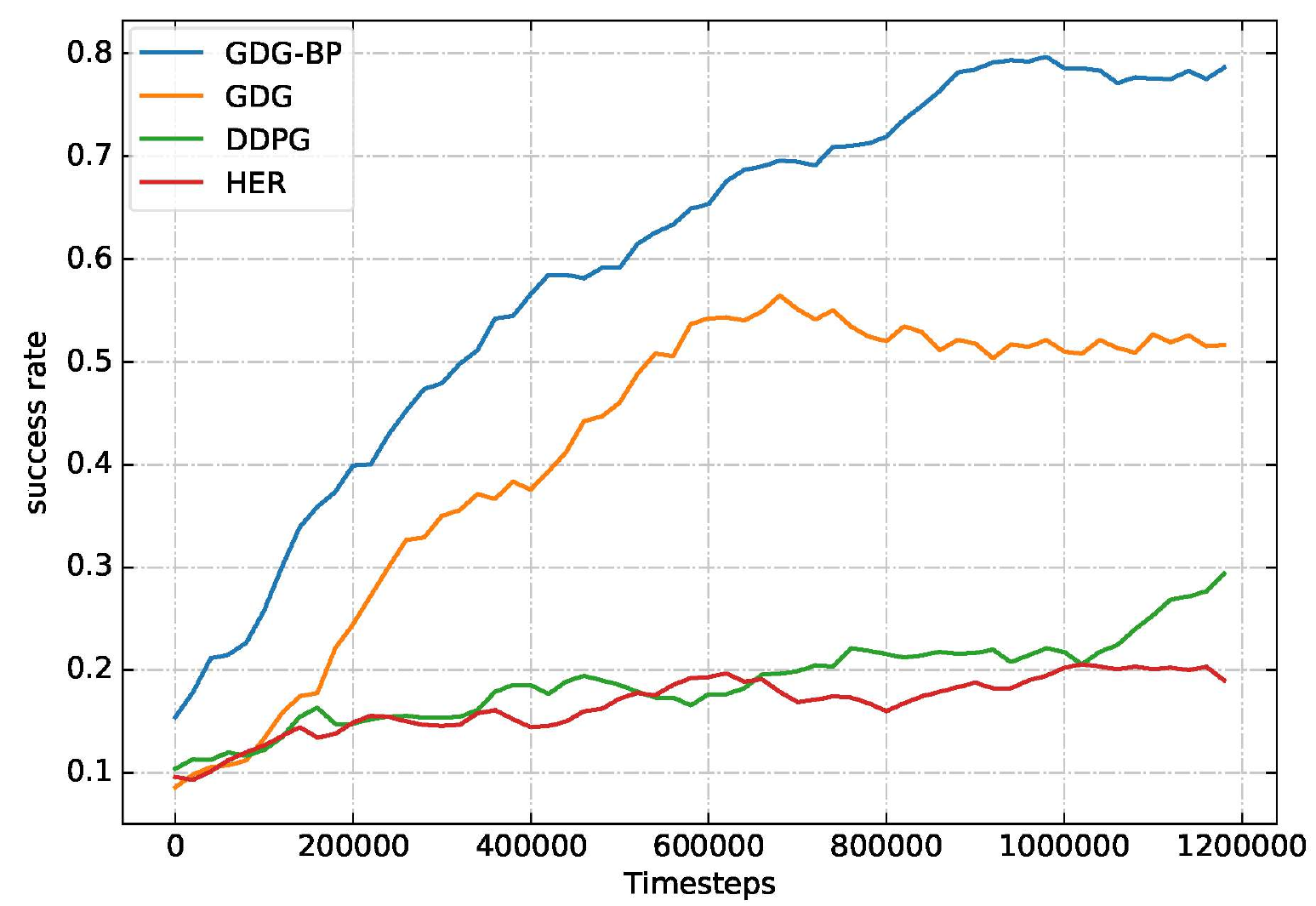}
\label{3}
\caption{Comparison of our method with DDGP and HER in the city environment over time with an average success rate.}
\end{figure}

\par We calculated the average success rate of each method at six different distances tasks in this environment. For each distance, we randomly generated 100 start and goal, and recorded whether each success. In each distance, if the goal is reached within 500 steps, it will be recorded as success, otherwise it will be recorded as failure. We repeated each experiment with five different random seeds. As shown in Figure 4, we plot the average results of five experiments as solid lines, and use translucent areas to represent the upper and lower limits of the five results. We can see the GDG with bridging planning can still maintain a relatively high success rate at longer distances, while other methods can only complete tasks at short distances.
\begin{figure}[h]
\centering
\includegraphics[width=7cm]{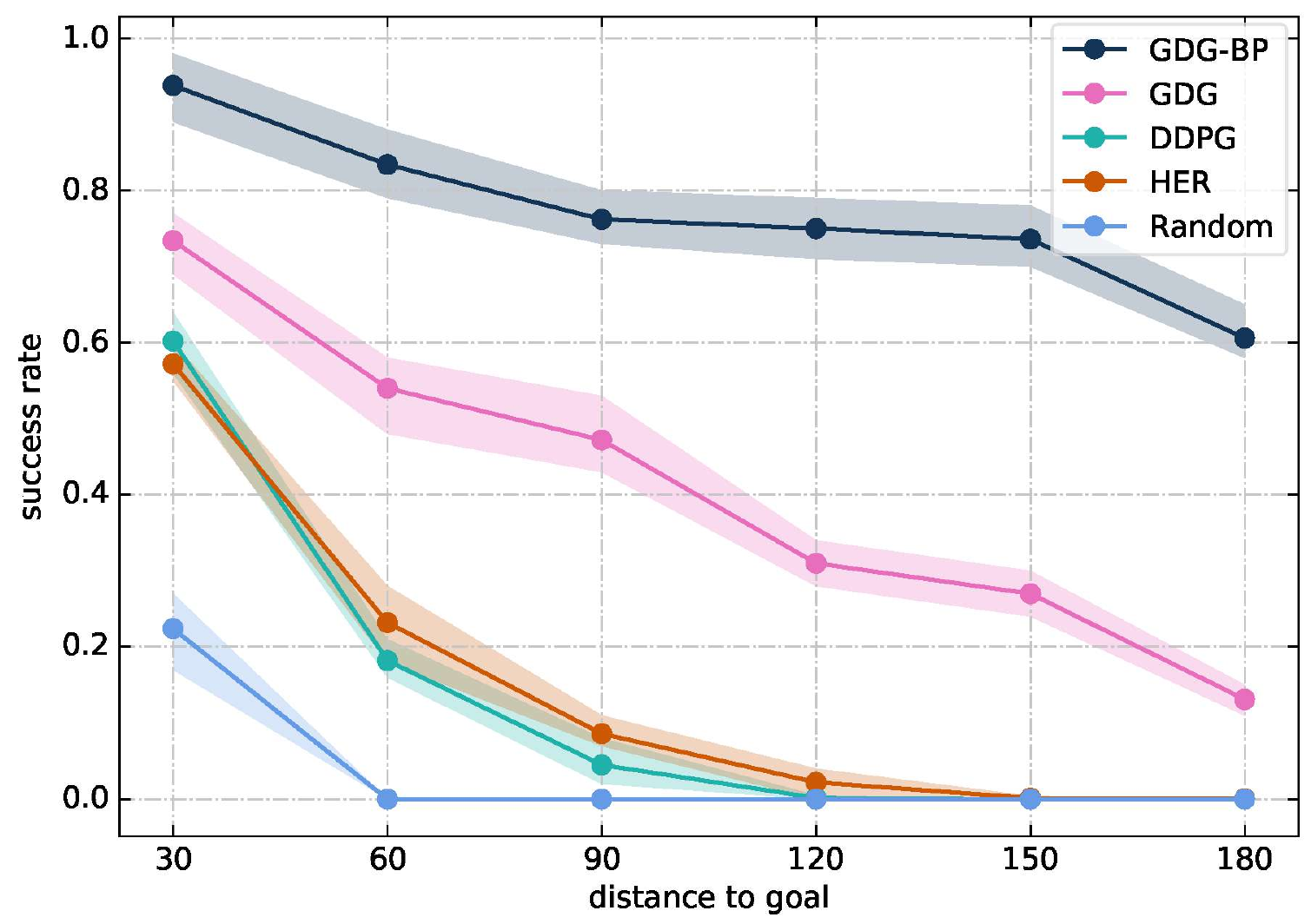}
\label{4}
\caption{Learning curves: The test is performed every 5 timesteps. The test uses the policy obtained to perform 100 times tasks, each task includes 50 different goals. Then use 5 different random seeds to repeat 5 times, and finally get the average success rate.}
\end{figure}

\par Figure 5 below shows the navigation results of our method at different distances between the start and the goal in the city environment. From the agent's path, it almost chooses a shortest distance to complete the task and achieve the goal as soon as possible.

\begin{figure}[htb]
\centering
{\includegraphics[width=8cm]{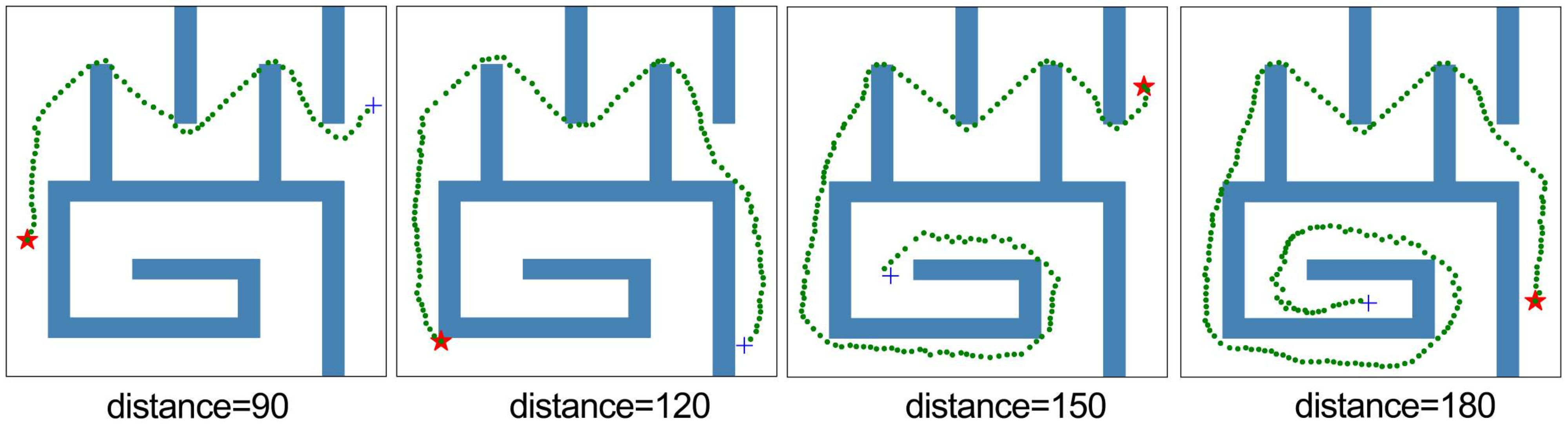}
\label{5}}
\caption{Performance of our method in different distances in city environment. From left to right, the results are displayed at distances of 90, 120, 150, and 180, respectively.}
\end{figure}

\subsection{Is GDG vulnerable to local optima if there is only one goal?}
We used a typical map with obstacles as shown in Figure 6 to evaluate the impact of local optima on our method and the general algorithm. The environment sets an area that is close to the goal but cannot reach the it as the local optimal solution. However, the actual optimal path needs to be far away from the local optimal area and get more negativ feedback before it reach the desired goal. Using distance as an evaluation indicator can easily cause the agent to fall into the local optimal area, because the local optimal area is close enough to the goal. At most 10 transfers are needed from the start to the local optimal area, and more than 80 transfers are needed to reach the desired goal. We evaluate four methods: the DDPG~\cite{lillicrap2016continuous} with the distance function as negative reward, the DDPG with the sparse reward, the GDG and the GDG with bridging planning. 
\begin{figure}[h]
\centering
{\includegraphics[width=8cm]{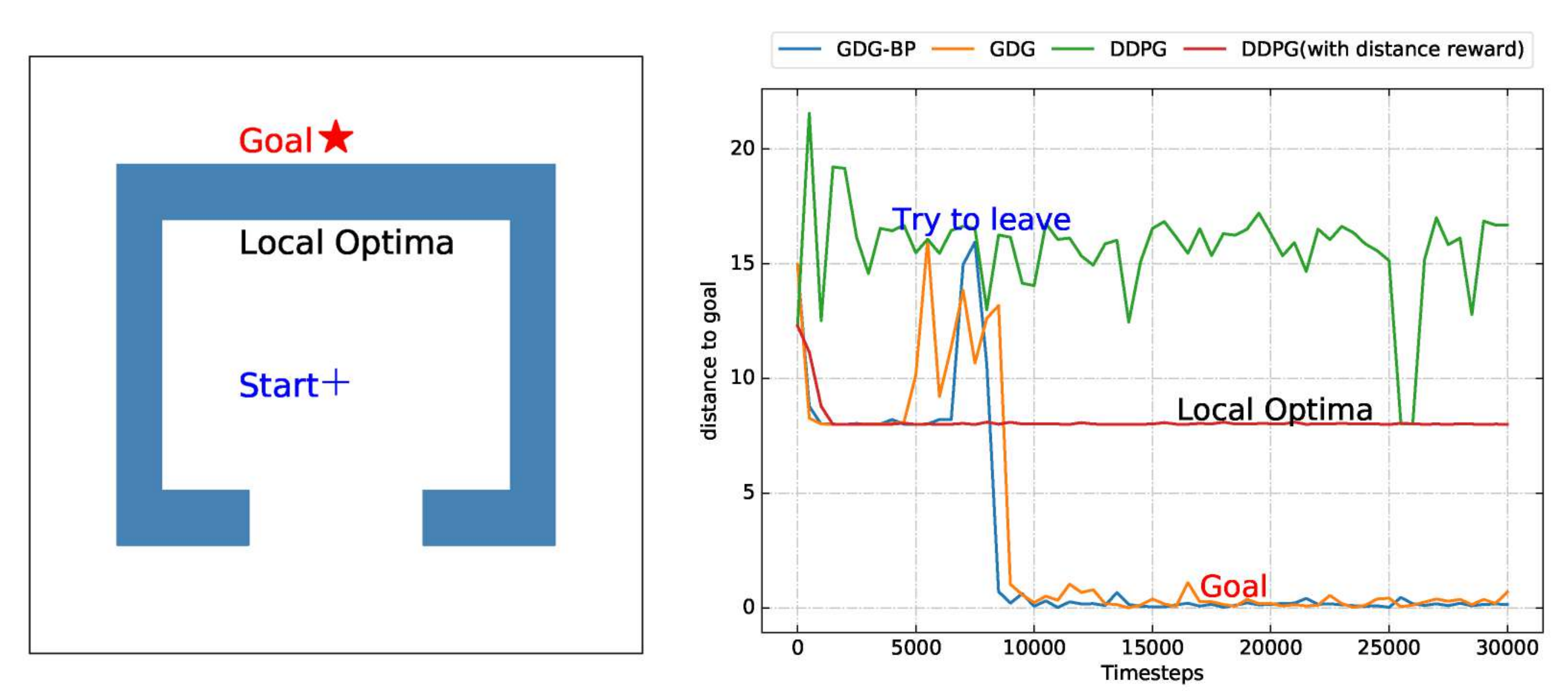}
\label{7}}
\caption{Left: environment with only one goal; Right: the final distance of the agent from the goal under four methods of training.}
\end{figure}

\par At the same exploration rate, experiments show that GDG and GDG with bridging planning can avoid the attraction of the local optimal area and reach the goal, while other two methods cannot complete the task. The performance of GDG and GDG with bridging planning in this experiment is consistent. By analyzing the distance curve on the right in Figure 6, we can see that in the initial training period, all methods except DDPG trappend in the local optimal area. However, in subsequent training, GDG and GDG with bridging planning trying to leave this. In the end they found the right way to reach the goal. The DDDPG with distance reward also falls into the local optimal area after training, but it has been trapped in the area since then, unable to reach the target. The DDPG that can only get rewards after reaching the goal is always exploring, and the minimum distance from the goal is in the local optimal area.

\section{Conclusions}
In this paper, We introduce a general method to solve the sparse reward and non-reward task by distance evaluation. We further propose a gradient algorithm based on goal-distance to solve the question that how distance evaluation can be used to improve policy. In task where the actual distance can not be calculated in advance, the distance estimation is used to assist the bridge planning to increase the exploration rate. The experimental results show, our method performs better than previous algorithms in complex sparse reward tasks, and can avoid the impact of local optimum on learning effectiveness. As future work, We hope that the agent can learn to set the goal it wants to reach to explore the unknown environment and decompose complex problems just like finding bridge points.

\bibliographystyle{named}
\bibliography{start}

\end{document}